\documentclass[10pt]{article}

\usepackage[utf8]{inputenc}
\usepackage[T1]{fontenc}
\usepackage{lmodern}
\usepackage{siunitx}
\usepackage[a4paper,margin=1in]{geometry}

\usepackage{graphicx}
\usepackage{caption}
\usepackage{subcaption}
\usepackage{pifont}
\usepackage{fontawesome5}
\usepackage{amsmath,amssymb,amsfonts}
\usepackage{bm} 

\usepackage{algorithmic}

\usepackage[l3]{csvsimple}

\usepackage{pgf}
\usepackage{pgfmath}
\usepackage{pgfplots}
\pgfplotsset{compat=1.18}
\usepackage{pgffor}
\usepackage{tikz}
\usepackage{printlen}
\usepackage{tabularray}
\usepackage{multirow, makecell, booktabs}
\usepackage[table]{xcolor}
\usepackage{collcell}
\usepackage{tabularx}
\usepackage{xcolor}
\usepackage{colortbl}
\usepackage{threeparttable}  
\definecolor{headerblue}{RGB}{173, 216, 230}  

\usepackage[authoryear]{natbib}

\usepackage{titlesec}
\titleformat{\section}{\large\bfseries}{\thesection}{1em}{}
\titleformat{\subsection}{\normalsize\bfseries}{\thesubsection}{1em}{}

\usepackage{url}
\usepackage[hidelinks]{hyperref}

\title{Self-supervised Pre-training Helps Retinal Disease Progression Modelling Most When Data Is Scarce}
\author{
Ifeoma Veronica Nwabufo\textsuperscript{1,2} 
\and
Julius Gervelmeyer\textsuperscript{1,2}
\and
Sarah Müller\textsuperscript{1,2}
\and
Philipp Berens \textsuperscript{1,2}\thanks{Corresponding author: \texttt{philipp.berens@uni-tuebingen.de}}
}

\date{}
\begin{document}
\maketitle

\noindent
\textsuperscript{1} Hertie Institute for AI in Brain Health, Faculty of Medicine, University of Tübingen, Germany\\
\textsuperscript{2} Tübingen AI Center, University of Tübingen, Germany

\begin{abstract}
    \noindent Modelling how a disease progresses over time requires longitudinal imaging cohorts, which are scarce and small, whereas cross-sectional data -- one image per participant -- is abundant. Self-supervised pre-training on such data offers a way to bridge this gap, but it is unclear which strategy best supports progression modelling, or how that answer depends on the amount of labelled longitudinal data. We study this for age-related macular degeneration (AMD), pre-training encoders on the large cross-sectional NAKO cohort and predicting time to late AMD on the longitudinal AREDS dataset. We compare in-house self-supervised encoders against a general-purpose (DINOv2) and a domain-specific (RETFound) foundation model, across contrastive, masked-autoencoding, and self-distillation objectives, under frozen and fine-tuned protocols, and across labelled training sets from 100 to 32,250 examples. Which model performs best depends on how the encoder is used. When the encoder is frozen and labels are few -- the regime typical of longitudinal cohorts -- pre-trained representations reach clinically reasonable discrimination from a few hundred labelled samples, while models trained from scratch do not; this advantage fades under fine-tuning. Transfer is governed by the self-supervision objective rather than corpus scale or domain match, so that an encoder pre-trained on a modest cross-sectional cohort matches or exceeds a far larger in-domain foundation model. Together, these results offer a practical recipe for building progression models where longitudinal data is scarce: a frozen self-supervised encoder with a lightweight survival head.
\end{abstract}

\section{Introduction}

In medical imaging, modelling how a disease progresses over time, rather than merely whether it is present at a single visit, requires longitudinal cohorts in which the same patients are imaged repeatedly over months or years. Yet such cohorts are scarce and expensive, because following and re-imaging the same individuals is slow and costly, and the labelled longitudinal datasets that reach public use are often correspondingly small. The widely used OASIS progression cohorts, for example, comprise a few hundred to about a thousand subjects~\citep{marcus2010open}, and even large screening trials such as the National Lung Screening Trial yield only a few hundred participants once one restricts to those with usable labelled follow-up imaging~\citep{nlst2011}. The Alzheimer's Disease Neuroimaging Initiative (ADNI) follows on the order of a thousand participants with serial MRI and PET~\citep{jack2008longitudinal} but remains the exception. Cross-sectional data, by contrast, is abundant: population-scale resources such as the UK Biobank image each of a hundred thousand participants essentially once~\citep{littlejohns2020uk}. This asymmetry -- abundant single-visit snapshots, scarce repeated observations -- is a central obstacle for disease progression modelling, and it compounds an already severe shortage of labelled medical images relative to natural images~\citep{varoquaux2022machine}.

Ophthalmology exemplifies this problem but also provides the data to study it directly. Many retinal diseases unfold over years, so prevention requires knowing not just whether a disease is present but when it will progress. This is a problem more naturally cast as survival analysis than classification, and one that depends on longitudinal data. Much of the AI work in ophthalmology has instead targeted disease detection~\citep{ting2017development, gulshan2016development, de2018clinically}, with progression modelling receiving comparatively little attention. Some studies have addressed it for age-related macular degeneration (AMD), a retinal disease that is among the leading causes of central vision loss in the elderly~\citep{burtoneyehealth}: \citet{yan2020deep} predicted time to late AMD from fundus photographs and genotypes on the AREDS cohort, \citet{peng2020predicting} paired image classification with survival analysis to match retinal specialists' prognostic accuracy, and \citet{rivail2023deep} showed that survival modelling of longitudinal OCT scans outperforms classification for AMD risk prediction. More recently,~\citet{gervelmeyer2024interpretable} modelled AMD progression with an interpretable-by-design survival network. What makes this line of work possible are datasets like the Age-Related Eye Disease Study (AREDS), which followed thousands of eyes over more than a decade~\citep{age1999age}, giving ophthalmology a large longitudinal cohort of the kind many fields lack.

In principle, the abundant cross-sectional data could offer a way around the scarcity of longitudinal labels. Self-supervised learning (SSL) can learn disease-relevant features from unlabelled images and transfer them to a smaller labelled task~\citep{shurrab2022self, huang2023self}, and three broad routes exist for obtaining such a representation in ophthalmology. General-purpose foundation models pre-trained on natural images (e.g.\ DINOv2) are cheap to reuse, but the benefit of natural-image pre-training for medical tasks is inconsistent and depends on task and training regime~\citep{valverde2026fine}. Domain-specific medical models, pre-trained directly on large unlabelled retinal cohorts (e.g.,\ RETFound \citep{zhou_foundation_2023}), are assumed to transfer better, though the evidence comes from classification rather than progression. A third route, pre-training one's own encoder on an in-house or publicly available cross-sectional cohort, exploits the abundant snapshot data directly and allows for full control over the training data: \citet{zhou2026understanding} compared pre-training data across two large fundus cohorts, but evaluated only on classification tasks. These three routes have not been compared for survival prediction, and it remains unknown which best supports progression modelling, or how that answer changes as labelled longitudinal data becomes scarce.

Here, we study these questions directly through a systematic evaluation of pre-training strategies for deep survival analysis, isolating which properties of a pre-trained encoder govern how well it transfers. We pre-train encoders on the large cross-sectional NAKO Ophthalmology dataset~\citep{Roa2026.05.04.26352019} and evaluate time-to-late-AMD prediction on the longitudinal AREDS dataset, comparing in-house SSL encoders against a general-purpose (DINOv2 \citep{oquab2024dinov}) and a domain-specific (RETFound \citep{zhou_foundation_2023}) foundation model, across contrastive, masked-autoencoding, and self-distillation objectives, and under both frozen and fine-tuned protocols. Because AREDS is large enough to subsample, we vary the labelled training set from 100 to the full 32,250 examples and measure directly how progression modelling degrades as longitudinal labels become scarce, an analysis that smaller longitudinal cohorts do not permit. Rather than only reporting which model ranks highest, we ask which factor -- the self-supervision objective, the pre-training corpus scale, the domain match, or the evaluation protocol -- accounts for the differences we observe, and find that the objective dominates: it explains essentially all of the variation between encoders, concentrated in the frozen, low-data regime, whereas neither corpus scale nor domain match adds much. This yields a practical and inexpensive recipe for the data-scarce setting typical of longitudinal cohorts: a frozen self-supervised encoder with a lightweight survival head.

\section{Methods}

\begin{figure}[!ht]
\centering
\includegraphics{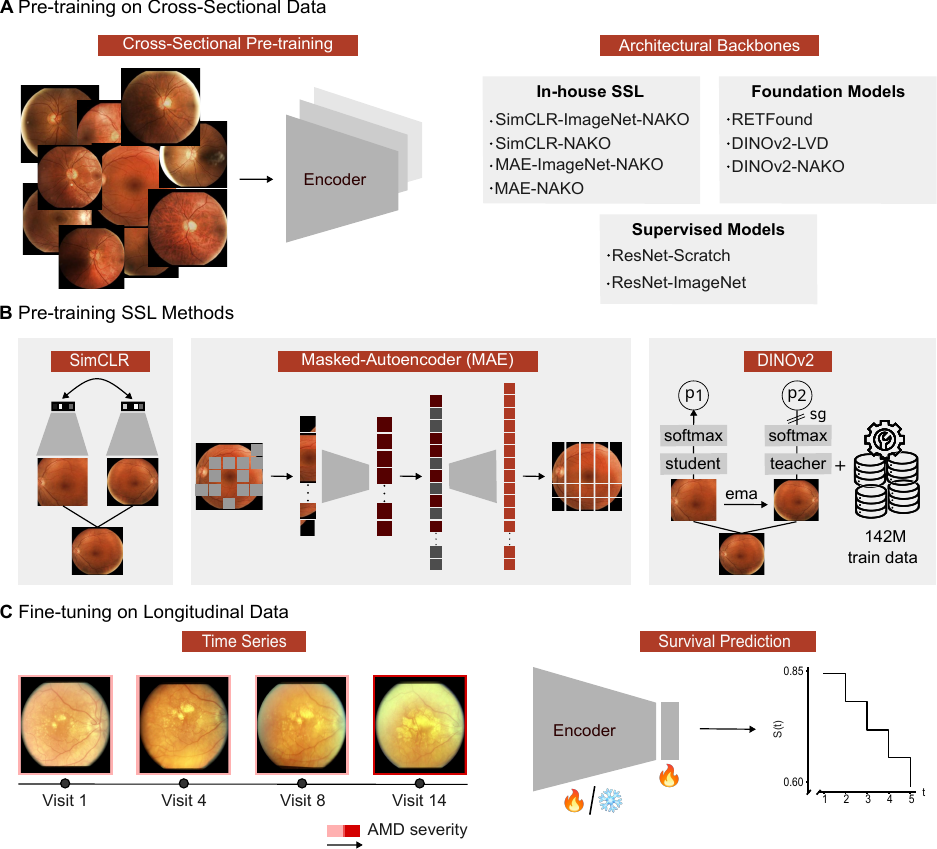}
\caption{Overview of the study framework. (\textbf{A}) Cross-sectional fundus images from the NAKO study are used for self-supervised pre-training with (\textbf{B}) three strategies: SimCLR, MAE, or DINOv2. (\textbf{C}) Downstream evaluation on longitudinal fundus image sequences from AREDS, here showing visits 1, 4, 8, and 14 for an example patient progressing from intermediate (AMD severity 9) to late AMD (severity 10). The resulting pre-trained encoder, either frozen or fine-tuned on AREDS, feeds a survival head, which generates predicted survival curves, $S(t)$, giving the probability of remaining free from late AMD over the following 1--5 years.}
\label{fig:main_figure}
\end{figure}

We used a two-stage framework to predict the time to conversion to late age-related macular degeneration (AMD) from fundus imaging (Figure~\ref{fig:main_figure}). We first pre-trained an encoder on cross-sectional data, then fine-tuned or froze it, before pairing it with a survival head to predict AMD progression on a longitudinal dataset.

\subsection{Datasets}

We used two datasets: the cross-sectional German National Cohort (NAKO)~\citep{nako, Roa2026.05.04.26352019} for pre-training, and the longitudinal Age-Related Eye Disease Study (AREDS)~\citep{age1999age} for the downstream survival task (Table~\ref{tab:dataset_des}).

NAKO is Germany's largest long-term population study of common diseases, with baseline examinations carried out between 2014 and 2019 and follow-up visits thereafter~\citep{nako, Roa2026.05.04.26352019} (Figure~\ref{fig:main_figure}A). More than 205,000 participants aged 20 to 69 were enrolled, with 48,460 participants receiving retinal imaging in the baseline visit. Most eyes were imaged in two different configurations (macula-centred, optic-disc-centred), and we used all available configurations per eye. We filtered for good quality using the Fundus Image Toolbox  (FIT)~\citep{gervelmeyer2025fundus}, resulting in 153,134 images, which we used for pre-training. The eye diseases in the dataset were self-reported: 5,102 participants reported having an eye disease, and 382 of these self-reported having AMD.

AREDS is a longitudinal study of 4,757 participants aged 55 to 80, followed over 12 years~\citep{age1999age}. Each retinal scan comes with an AMD severity label that ranges from 1 to 12, where 1 to 9 mark increasing risk of conversion to late AMD and 10 to 12 denote late AMD, based on drusen and other macular characteristics. Its long follow-up makes AREDS suitable for a survival task, with repeated gradings per eye over time. After removing duplicates and low-quality images, we retained a single orientation for eyes photographed from multiple angles (stereoscopic view selected at random). We treated the left and right eyes of the remaining 4,426 participants as independent observations since each eye could progress differently. This resulted in 8,784 eyes with multiple visits, yielding 55,173 images in total. 1,988 eyes were diagnosed with late AMD throughout the course of the study. We split the dataset by patient into training, validation, and test sets using a 60/20/20 ratio, ensuring each patient appeared in only one split.

\begin{table}[!t]
\centering
\caption{\textbf{Dataset characteristics}}
\label{tab:dataset_des}
\begin{threeparttable}
\renewcommand{\arraystretch}{1.15}
\begin{tabular}{@{}l S[table-format=6.0] S[table-format=6.0]@{}}
\toprule
 & {\textbf{NAKO}} & {\textbf{AREDS}} \\
\midrule
\multicolumn{3}{@{}l}{\textit{Cohort size}} \\
\quad Images, \textit{n}       & {153,134} & {55,173} \\
\quad Eyes, \textit{n}         & {84,723}  & {8,784}  \\
\quad Visits, \textit{n} mean $\pm$ \textit{SD} & {1.0}  & {$12.5 \pm 6.3$} \\
\addlinespace
\multicolumn{3}{@{}l}{\textit{Demographics}} \\
\quad Male, \textit{n}             & {25,534}  & {1,943}   \\
\quad Female, \textit{n}           & {22,926}  & {2,483}   \\
\quad Age, y            & {$48.9 \pm 12.5$} & {$74.2 \pm 5.5$} \\
\addlinespace
\multicolumn{3}{@{}l}{\textit{Clinical}} \\
\quad Labels used, \textit{n}    & {--} & {12} \\
\quad AMD & {382\tnote{a}} & {1,988\tnote{b}} \\

\bottomrule
\end{tabular}
\begin{tablenotes}
  \footnotesize
  \item[a] Individuals who self-reported that they had AMD.
  \item[b] Value reflects those with late AMD.
\end{tablenotes}
\end{threeparttable}
\end{table}

\subsection{Encoder Models \& Pre-training}

We evaluated different encoder models and different pre-training strategies regarding their usefulness for longitudinal disease modelling (Figure \ref{fig:main_figure} and Table \ref{tab:summary_of_encoders}).

\paragraph{Foundation models.} These models are typically trained using a two-stage paradigm~\citep{zhou2023foundation, oquab2024dinov}: pre-training on large unlabelled datasets, then fine-tuning on smaller labelled ones. We evaluated two models of this type: (i) RETFound, an ophthalmological foundation model with a reconstruction objective~\citep{he2022masked}, for which we used its colour fundus photography variant, pre-trained on over 900{,}000 fundus images, initialised from ImageNet weights~\citep{zhou2023foundation}; and (ii) DINOv2~\citep{oquab2024dinov}, a general-purpose model pre-trained on a large vision dataset of 142 million natural images (LVD-142M) with a self-distillation objective~\citep{hinton2015distilling}, which performs well across downstream tasks without fine-tuning.

\paragraph{In-house models.} Alongside the foundation models, we pre-trained our own encoders on the NAKO dataset using three SSL methods -- SimCLR~\citep{chen2020simple}, Masked Autoencoder (MAE)~\citep{he2022masked}, and DINOv2~\citep{oquab2024dinov} -- each adapted to retinal imaging as described below.

\begin{itemize}
    \item SimCLR is a contrastive learning (CL) method that pulls representations of augmented views of the same image together and pushes different images apart~\citep{chen2020simple}, so the choice of augmentations drives downstream performance. We followed the original augmentation pipeline and added rotations, which improve downstream performance on medical images~\citep{nwabufo2024selfsupervised}. We used a ResNet18~\citep{he2016deep} backbone with two initialisations (random weights and ImageNet), a batch size of 256 and pre-trained for 200 epochs.
    
    \item We pre-trained MAEs following RETFound's training pipeline~\citep{he2022masked,zhou2023foundation}. We additionally applied the global/local crop idea from DINOv2~\citep{oquab2024dinov}, and treated the full retinal image as a global view and the macular region as a local view, doubling the effective training set. We used a 50\% masking ratio rather than the more standard higher ratio, since the region of interest in medical images is localised and benefits from more context during reconstruction~\citep{xie2024rethinking}. We used a ViT-Base~\citep{dosovitskiy2020image} backbone with two initialisations (random weights and ImageNet weights). We pre-trained for 300 epochs, with an image size of $224\times 224$ and a patch size of 16. 

    \item Because DINOv2 performs well out of the box, we adapted it to retinal images through parameter-efficient fine-tuning with LoRA~\citep{lora, zhu_melo_2024} rather than pre-training from scratch, which would have been infeasible given the number of parameters. We initialised both teacher and student from DINOv2 weights, froze the student, and attached learnable adapters to the attention layers, leaving few trainable parameters. As our dataset is far smaller than the original DINOv2 corpus, we reduced the number of prototypes to 16{,}384.
\end{itemize}

\paragraph{Supervised Baselines.} To have a baseline comparison for the downstream survival prediction that did not employ SSL pre-training, we used ResNet18~\citep{he2016deep} as a baseline with two variants: ResNet18 with random weights and ResNet18 with ImageNet weights, each evaluated only on the AREDS dataset.

\subsection{Survival Modelling}

Survival models predict the time until an event of interest occurs, the \textit{time-to-event}. While originally developed to model time until death~\citep{chen2024introductiondeepsurvivalanalysis}, here we use them to model the time until conversion to late AMD. What distinguishes survival analysis from standard regression or classification is \textit{censoring}: for some subjects, the event is not observed within the study period -- because the study ends, the subject drops out, or an unrelated event intervenes -- yet these subjects still carry information, since we know the event had not occurred by their last observation. We encode each subject as a tuple $(\delta, T, X)$, where $T$ is the time until the event or censoring for input $x \in X$ and the indicator $\delta$ is $1$ if the event was observed and $0$ if the subject was censored. We treat each scan as a subject.

We characterise risk through the survival function $S(t \mid x) = 1 - F(t \mid x)$, the probability of remaining event-free beyond time $t$, where $F(t \mid x) = \mathbb{P}(T \leq t \mid X = x)$ is the cumulative distribution function. The related hazard function
\begin{equation}
    h(t \mid x) = \frac{f(t \mid x)}{S(t \mid x)},
\end{equation}
gives the instantaneous event risk at time $t$ given survival up to $t$. Integrating it yields the cumulative hazard $H(t \mid x) = \int_0^t h(u \mid x)\,du$, from which the survival function follows as $S(t \mid x) = e^{-H(t \mid x)}$.

We model the hazard with the \textit{Cox proportional hazards model}~\citep{cox1972regression, chen2024introductiondeepsurvivalanalysis}, which factorises it into a shared baseline hazard function $h_0(t)$ and a subject-specific, time-independent risk score $G(x)$:
\begin{equation}
     h(t \mid x) = h_{0}(t)\, e^{G(x)}, \quad t \ge 0,\ x \in X.
\end{equation}
As the Cox model cannot handle high-dimensional inputs, such as fundus images, we parametrise $G(x)$ using a neural network with a single output node and estimate the baseline hazard $h_0(t)$ using the non-parametric Breslow estimator~\citep{breslow1972discussion}. Integrating the fitted hazard over time and substituting into $S(t \mid x)$ recovers the full survival curve for each subject.

To receive the literal time-to-event prediction, simple probability thresholding can be applied to the survival curve. Since this would collapse the curve to one value with practical importance but fewer dimensions, we compared models based on the full predicted survival curves and risk logits.

\begin{table}[!t]
    \centering
    \caption{Summary of encoders used in this study. \ding{51}/\ding{55} under \emph{In-domain} indicates whether the pre-trained data contained fundus images.}
    \label{tab:summary_of_encoders}
    \resizebox{\textwidth}{!}{%
    \begin{tabular}{@{}ccccccc@{}}
        \toprule
        \# & \makecell{Pretrain\\Method} & \makecell{Pretrain\\Objective} & \makecell{Pretrain\\Images} & In-domain & Family & Model Name \\
        \midrule
        \multicolumn{7}{@{}l}{\textit{Baselines (no self-supervised pretraining)}} \\
        1 & -- & -- & -- & \ding{55} & Supervised & ResNet-Scratch \\
        2 & -- & \makecell{Supervised\\classification} & ImageNet & \ding{55} & Supervised & ResNet-ImageNet \\
         & & & & & & \\

        \multicolumn{7}{@{}l}{\textit{Contrastive learning}} \\
        3 & SimCLR & Contrastive & NAKO & \ding{55} & SSL & SimCLR-NAKO \\
        4 & SimCLR & Contrastive & \makecell{ImageNet +\\NAKO} & \ding{51} & SSL & SimCLR-ImageNet-NAKO \\
         & & & & & & \\

        \multicolumn{7}{@{}l}{\textit{Masked image reconstruction}} \\
        5 & MAE & Reconstruction & \makecell{ImageNet +\\Moorfields} & \ding{51} & FM & RETFound \\
         & & & & & & \\
        6 & MAE & Reconstruction & NAKO & \ding{51} & SSL & MAE-NAKO \\
         & & & & & & \\
        7 & MAE & Reconstruction & \makecell{ImageNet +\\NAKO} & \ding{51} & SSL & MAE-ImageNet-NAKO \\
         & & & & & & \\

        \multicolumn{7}{@{}l}{\textit{Self-distillation}} \\
        8 & DINOv2 & Self-distillation & LVD-142M & \ding{55} & FM & DINOv2-LVD \\
        9& DINOv2 + LoRA & Self-distillation & \makecell{LVD-142M +\\NAKO} & \ding{51} & FM & DINOv2-NAKO \\
         & & & & & & \\
        \bottomrule
    \end{tabular}%
    }
\end{table}

\subsection{Evaluation}
\label{subsec:evaluation}
We evaluated model performance along two dimensions, using survival analysis metrics rather than classification ones: discrimination, how well the model ranks subjects by risk, and calibration, how well predicted survival probabilities match observed outcomes~\citep{park2021review}.

For discrimination, we used Harrell's concordance index (C-index)~\citep{harrell1996multivariable, park2021review}, the fraction of comparable subject pairs (pairs where the observed data can determine which subject experienced the event first) whose predicted risk ordering agrees with their observed event-time ordering:
\begin{equation}
C = \frac{\sum_{i,j} \mathbf{1}(T_i < T_j)\, \mathbf{1}(\hat{\eta}_i > \hat{\eta}_j)\, \delta_i}{\sum_{i,j} \mathbf{1}(T_i < T_j)\, \delta_i},
\end{equation}
where $\hat{\eta}_i = G(x_i)$ is the risk score the model predicts for subject $i$. A pair is concordant when the higher-risk subject has the shorter time-to-event ($\hat{\eta}_i > \hat{\eta}_j$ and $T_i < T_j$). A C-index of 0.5 is random, and 1.0 is a perfect ranking.

For calibration, we used the Brier score (BS)~\citep{graf1999assessment}, which in its basic form is the squared difference between the predicted survival probability and the observed survival status at time $t$:
\begin{equation}
\text{BS}(t) = \frac{1}{N} \sum_{i=1}^{N} \left( \hat{S}_i(t) - \mathbf{1}(T_i > t) \right)^2,
\end{equation}
where $\hat{S}_i(t)$ is the predicted survival probability for subject $i$. We used the inverse-probability-of-censoring weighted variant as implemented in scikit-survival~\citep{sksurv}. To summarise calibration over time, we report the integrated Brier score (IBS)~\citep{graf1999assessment},
\begin{equation}
    \text{IBS} = \frac{1}{t_{\max}} \int_0^{t_{\max}} \text{BS}(t)\, dt,
\end{equation}
integrated over 1 to 5 years after the queried visit. A lower IBS indicates better calibration, with 0 a perfect prediction and 0.25 an uninformative model.  
 
\subsection{Experimental Details}
After pre-training, we froze or fully fine-tuned each encoder and paired it with either a linear layer or a two-layer MLP head, each yielding a single logit for the survival task. We trained each survival model for up to 50 epochs using the CoxPH loss~\citep{katzman2018deepsurv}, with early stopping after 20 epochs without improvement in validation IBS.  We used an image size of $224\times 224$ and applied crops, flips, colour jitter, rotation, and image normalization augmentations during training, but only crops and normalization during testing. We used 7 training sizes: 100, 500, 1,000, 10,000, 20,000, 30,000, and 32,250. We adjusted the batch size according to the training set size: 8 for training sets of 100 samples, 32 for training sets between 100 and 5,000 samples, and 64 for training sets of 10,000 samples or more in order to ensure sufficient gradient updates for smaller datasets while improving gradient estimate accuracy for larger ones. We used the AdamW optimiser with a learning rate of $1e^{-5}$ for the encoder and $4e^{-4}$ for the MLP, both with a weight decay of $1e^{-2}$. The MLP had a dropout of 0.1. For each training set size, we trained each model 5 times with different random seeds and evaluated it on a common, fixed held-out test set, reporting the mean and standard deviation across the 5 runs. Each model was trained with an NVIDIA GeForce RTX 2080 or A100 Ti GPU with the PyTorch framework~\citep{paszke2019pytorch}.

\subsection{Statistical analysis}

To identify which properties of an encoder drive survival prediction, we modelled the C-index and IBS as smooth functions of
training set size using generalised additive models (GAMs; \texttt{mgcv}~\citep{wood2011fast}). We averaged the five training runs within each combination of model, training size, survival head, and evaluation protocol, giving one value per cell. Since the five runs are repeated measurements of the same configuration, we averaged them to avoid pseudoreplication so that each cell acts as a single independent observation of an \textit{average run}. Therefore, no random effect for the replication structure was needed. In each model, a thin-plate regression spline of $\log$ training size was allowed to differ by the interaction of the pre-training objective (contrastive, reconstruction, self-distillation, supervised, or none), the evaluation protocol, and the survival head, with a matching parametric term for the level of each curve.

To quantify how much each factor explained, we compared nested models by the Akaike information criterion (AIC), all fitted by maximum likelihood so that
their fixed-effect structures were comparable. To assess the goodness of fit, we additionally report the percentage of deviance explained, which quantifies the proportion of total variation captured by the smooth terms. Starting from a baseline containing only training size, protocol, and head, we added the pre-training
objective and then the model family (in-house SSL, foundation model, or baseline) and lastly, the pre-training domain (in- or out-of-domain), recording
the change in AIC and in deviance explained  (Table~\ref{tab:aic_model_comparison}). We repeated the decomposition within each protocol--head combination to locate where each factor mattered because the objective factor is partly confounded with the pre-trained-versus-baseline contrast (Table~\ref{tab:regime_decomposition}). Next, we decomposed the results within the three self-supervision objectives alone (Table~\ref{tab:regime_decomposition_ssl_subset}).  Lastly, we summarise the influence of the objective as the range of predicted performance across objectives at each training size (the objective spread, Figure~\ref{fig:objective} \&~\ref{fig:objective_ibs}).

\begin{figure}[t!]
\centering
\includegraphics[scale=1]{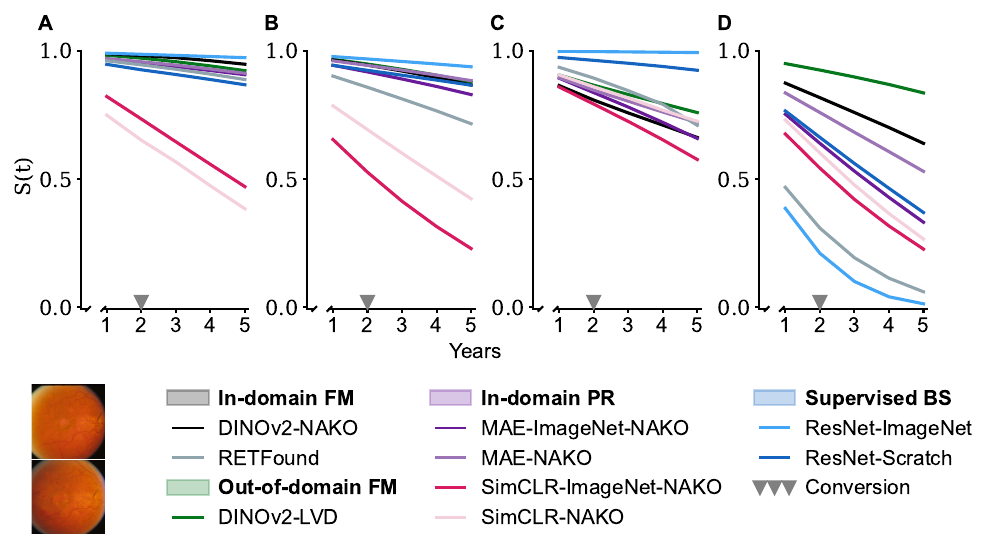}
\caption{Survival curves $S(t)$ reflecting the probability of remaining free from late AMD progression in the next 1--5 years, for one patient, based on the fundus image at year 2 (top fundus image), as predicted using different encoder models. The grey triangle marks the screening with observed conversion (fundus image shown at the bottom). (\textbf{A}, \textbf{B}) Survival model with frozen encoder, with read-out trained on (\textbf{A}) 1,000  and (\textbf{B}) 32,250  examples. (\textbf{C}, \textbf{D}) Survival model with fine-tuned encoder, trained on (\textbf{C}) 1,000  and (\textbf{D}) 32,250 examples. Colours indicate model family (see legend).}
\label{fig:survival_curves}
\end{figure}

\section{Results}

We asked whether large cross-sectional cohorts with tens of thousands of participants, in which each contributes roughly a single image to the dataset, can inform disease-progression models trained on the much smaller longitudinal cohorts that such prognostic tasks require. We studied this question for fundus imaging in ophthalmology, using two datasets (Table~\ref{tab:dataset_des}): the German National Cohort Ophthalmology dataset (NAKO)~\citep{Roa2026.05.04.26352019}, a large, high-quality cross-sectional dataset from a broadly sampled German population, and the AREDS dataset~\citep{age1999age}, a longitudinal study of age-related macular degeneration progression. We pre-trained encoders on NAKO and asked which training strategy best supports a deep survival model predicting the time to conversion to late AMD on AREDS, using downstream training sets ranging from 100 examples to the full 32,250 to simulate the data scarcity typical of longitudinal cohorts. These strategies spanned both encoders we pre-trained ourselves on NAKO using contrastive, reconstruction, and self-distillation objectives, and large publicly available foundation models applied off the shelf (RETFound, pre-trained on retinal images, and DINOv2, pre-trained on natural images). Across this framework, we then trained complete survival models consisting of an encoder, a read-out, and the survival loss function. We varied four factors that shape how well a representation transfers: the pre-training objective, the model family, the pre-training domain, and the evaluation protocol (frozen or fine-tuned; Table~\ref{tab:summary_of_encoders} \&~\ref{tab:models}), and asked which of them actually accounted for differences in performance.

For a single patient, the output of the model is a survival curve $S(t)$ reflecting the probability of remaining free from late AMD over the following one to five years based on a fundus image (Figure~\ref{fig:survival_curves}). We noticed that the survival curves predicted varied widely across models. In general, more training data led to more similar survival curves, indicating that different models became more similar as more training data was available (compare Figure~\ref{fig:survival_curves}A, C to B, D). Also, linear read-outs led to higher survival probabilities for many models, and they varied less over time (compare Figure~\ref{fig:survival_curves_linear}A, C to B, D).

\subsection{SimCLR-pretrained models had the best peak performance}
To quantify these effects, we assessed both how well the predicted survival probabilities matched observed outcomes (``calibration'', Integrated Brier Score, see \nameref{subsec:evaluation}) and how well the models ranked subjects by risk (``discrimination'', C-index). We first established which encoder performed best with abundant data. With a frozen encoder and an MLP head at $n = 32{,}250$, SimCLR-ImageNet-NAKO achieved the best discrimination and calibration (C-index $0.920 \pm 0.001$; IBS $0.054 \pm 0.001$; Table~\ref{tab:peak_performance}), followed closely by the two DINOv2 variants with C-Index $0.916 \pm 0.002$ and $0.912 \pm 0.003$. The MAE-based models reached lower peaks (RETFound $0.870$, MAE-ImageNet-NAKO $0.876$, MAE-NAKO $0.838$). Interestingly, RETFound performed similarly to our in-house trained SSL despite being pre-trained on a much larger corpus. However, the supervised baselines trailed every pre-trained model.

\begin{table}[t]
    \centering
    \caption{Performance at two training set sizes (\ding{55} = 1{,}000 samples, \ding{51} = full training set of 32{,}250) for a frozen encoder with an MLP head. Values are mean $\pm$ SD across five repeated training runs.}
    \label{tab:peak_performance}
    \begin{threeparttable}
    
    \setlength{\tabcolsep}{6pt}
    \begin{tabular}{lccc}
    \toprule
    \textbf{Model}& \makecell{\textbf{Train}\\\textbf{Size}} 
        & \textbf{C-index} $\uparrow$ 
        & \textbf{IBS} $\downarrow$  \\
    \midrule
    
    \multicolumn{4}{l}{\textit{SSL models}} \\
    
    \multirow{2}{*}{SimCLR-ImageNet-NAKO}  &  {\ding{55}} & \textbf{0.891 $\pm$ 0.002} & \textbf{0.063 $\pm$ 0.001} \\
        & {\ding{51}} & \textbf{0.920 $\pm$ 0.001} & \textbf{0.054 $\pm$ 0.001} \\
 
    \multirow{2}{*}{SimCLR-NAKO}   & {\ding{55}} & 0.879 $\pm$ 0.015 & 0.067 $\pm$ 0.003 \\
        & {\ding{51}} & 0.913 $\pm$ 0.009 & 0.057 $\pm$ 0.003 \\
   
\multirow{2}{*}{MAE-ImageNet-NAKO}  & {\ding{55}} & 0.809 $\pm$ 0.005 & 0.077 $\pm$ 0.001 \\
  & {\ding{51}} & 0.876 $\pm$ 0.008 & 0.064 $\pm$ 0.001 \\

    \multirow{2}{*}{MAE-NAKO}  & {\ding{55}} & 	0.769 $\pm$ 0.013 &0.084 $\pm$0.006 \\
  & {\ding{51}} & 0.838 $\pm$ 0.028 &0.069 $\pm$ 0.004 \\
    \midrule

    \multicolumn{4}{l}{\textit{Foundation models}} \\
    \multirow{2}{*}{DINOv2-LVD}      & {\ding{55}} & 0.885 $\pm$ 0.005 & 0.068 $\pm$ 0.004 \\
  & {\ding{51}} & 0.916 $\pm$ 0.002 & 0.057 $\pm$ 0.002 \\

    \multirow{2}{*}{DINOv2-NAKO} & {\ding{55}} & 0.889 $\pm$ 0.003 & 0.066 $\pm$ 0.004 \\
  & {\ding{51}} & 0.912 $\pm$ 0.003 & 0.057 $\pm$ 0.002 \\

    \multirow{2}{*}{RETFound}  
   & {\ding{55}} & 0.797 $\pm$ 0.020 & 0.080 $\pm$ 0.008 \\
  & {\ding{51}} & 0.870 $\pm$ 0.007 & 0.063 $\pm$ 0.001 \\
    \midrule
    
    \multicolumn{4}{l}{\textit{Supervised baselines}} \\
    \multirow{2}{*}{ResNet-ImageNet} 
     & {\ding{55}} & 0.737 $\pm$ 0.004 & 0.088 $\pm$ 0.005 \\
  & {\ding{51}} & 0.805 $\pm$ 0.008 & 0.075 $\pm$ 0.003 \\
    
    \multirow{2}{*}{ResNet-Scratch}  
   & {\ding{55}} & 0.648 $\pm$ 0.036 & 0.087 $\pm$ 0.001 \\
  & {\ding{51}} & 0.647 $\pm$ 0.038 & 0.087 $\pm$ 0.001 \\
 
    \bottomrule
    \end{tabular}
    \end{threeparttable}
\end{table}

\subsection{Pre-training benefits are most pronounced in the low-data regime}
\begin{figure}[t!]
\centering
\includegraphics{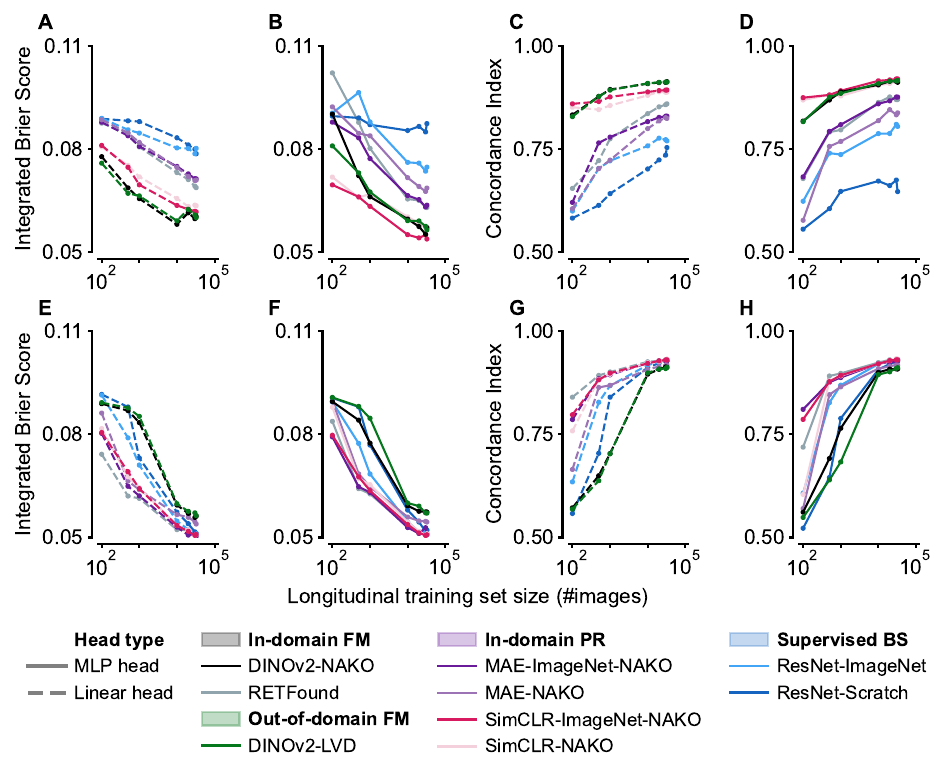}
\caption{Performance of frozen (\textbf{A}-\textbf{D}) and fine-tuned (\textbf{E}-\textbf{H}) encoder models predicting the probability of progression to late AMD in the next 1--5 years, as a function of training set size from 100 examples to full training data. Models were evaluated using the Integrated Brier Score (\textbf{A}, \textbf{B}, \textbf{E}, \textbf{F}; lower is better) and C-Index (\textbf{C}, \textbf{D}, \textbf{G}, \textbf{H}; higher is better), with a linear readout (\textbf{A}, \textbf{C}, \textbf{E}, \textbf{G}) or an MLP prediction head (\textbf{B}, \textbf{D}, \textbf{F}, \textbf{H}).}
\label{fig:frozen_encoder_full}
\end{figure}

Because longitudinal cohorts are typically small -- often falling toward the lower end of the size range we test here -- we examined each model's performance across training sizes from 100 to 32,250 examples, under both frozen and fine-tuned protocols (Figure~\ref{fig:frozen_encoder_full}).  Under a frozen encoder (Figure~\ref{fig:frozen_encoder_full}, top), the models differed widely in their ability to support good survival predictions: many pre-trained encoders held high calibration and discrimination down to a few hundred examples, while the supervised baselines degraded sharply with smaller data size. DINOv2 models and SimCLR-trained encoders performed especially well on small longitudinal datasets, while MAE-based encoders performed worse. Under fine-tuning, the models converged into a narrower band across most of the range, with the supervised baselines closing almost the entire gap for sufficiently large longitudinal datasets (Figure~\ref{fig:frozen_encoder_full}, bottom). In the fine-tuned case, SimCLR-trained and MAE-based encoders performed best. The benefit of pre-training therefore mattered mainly for the frozen encoder in the low-data regime and was largely absent for end-to-end fine-tuning on large longitudinal datasets.

\subsection{Minimum training set size for clinically reasonable performance}
To quantify this regime dependence, we determined the minimum training set size at which each model reached reasonable discrimination, using two criteria: an absolute threshold (C-Index $\geq 0.75$) and a relative threshold (within 95\% of the model's own peak at any training size). For each model and training size, we computed a bootstrap 95\% confidence interval (CI) for the mean C-Index (resampling the five training runs' predictions, 2{,}000 replicates) and took the smallest $n$ at which the lower bound reached the thresholds.

With the frozen encoder (Figure~\ref{fig:frozen_encoder_full}, top and Table~\ref{tab:min_training_size_frozen}), DINOv2 and SimCLR-based models reached both thresholds with very little data ($n \leq 100$ absolute; $n \leq 1{,}000$ relative), while the MAE-based models needed more data ($n \leq 1,000$ absolute; $n \leq 10{,}000$ relative). The supervised baselines showed worse performance as ResNet-ImageNet required $n \leq 10{,}000$, and ResNet-Scratch never reached a C-Index of $0.75$ within the evaluated range. Frozen pre-trained representations thus reached reasonable discrimination from a few hundred labelled samples, whereas a model without any prior pre-training did not reach comparable discrimination even with tens of thousands. 

Under fine-tuning (Figure~\ref{fig:frozen_encoder_full}, bottom and Table~\ref{tab:min_training_size_finetuned}), every model reached the absolute threshold by $n \leq 10{,}000$, and peak performance converged into a narrow range (between $0.908$ and $0.933$), against the much wider spread when frozen (range between $0.675$ and $0.920$). ResNet-Scratch changed most, from the weakest model when frozen (never reaching the absolute threshold) to among the strongest after fine-tuning ($0.675 \rightarrow 0.927$). The two DINOv2 variants moved the opposite way, from among the best when frozen to the lowest of the converged band after fine-tuning ($0.916 \rightarrow 0.911$; $0.914 \rightarrow 0.908$), and were the slowest to reach the threshold, consistent with prior evidence that DINOv2 is better used off-the-shelf~\citep{oquab_dinov2_2024}. Although the models converged to similar performance once enough data was available for fine-tuning (and thus the initial pre-training became less important), the models differed in their ability to reach the defined thresholds, with the MAE and SimCLR models reaching the absolute threshold fastest ($n \leq 500$), ahead of the DINOv2 models.

\begin{table}[t]
    \centering
    \caption{Minimum training set size for clinically reasonable performance (frozen encoder,
             MLP head). Absolute threshold: bootstrap 95\% CI lower bound of C-index $\ge 0.75$.  Relative threshold: CI lower bound $\ge$ 95\% of peak mean C-index.
    }
    \label{tab:min_training_size_frozen}
    \begin{tabular}{lccc}
    \toprule
    \textbf{Model} 
        & \makecell{\textbf{Min }$n$\\(CI $\geq 0.75$)}
        & \makecell{\textbf{Min }$n$\\(CI $\geq 95\%$ peak)}
        & {Peak C-index} [95\% CI]  \\
    \midrule
    \multicolumn{4}{l}{\textit{SSL models}} \\
    SimCLR-NAKO        & $\leq 100$    & $\leq 500$     & $0.913\ [0.906, 0.918]$ \\
    SimCLR-ImageNet-NAKO   & $\leq 100$    & $\leq 500$     & $0.920\ [0.919, 0.921]$ \\
    MAE-NAKO           &         $\le 1,000$          &         $\le 10{,}000$  & $0.846 \ [0.833, 0.856]$ \\
    MAE-ImageNet-NAKO     &         $\le 500$          &         $\le 10{,}000$  & $0.876 \ [0.876, 0.877]$ \\
    \midrule
    \multicolumn{4}{l}{\textit{Foundation models}} \\
    DINOv2-LVD        & $\leq 100$    & $\leq 1{,}000$ & $0.916\ [0.915, 0.917]$ \\
    DINOv2-NAKO         & $\leq 100$    & $\leq 1{,}000$ & $0.914\ [0.912, 0.916]$ \\
    RETFound            & $\leq 500$    & $\leq 10{,}000$& $0.877\ [0.872, 0.881]$ \\
    \midrule
    \multicolumn{4}{l}{\textit{Supervised baselines}} \\
    ResNet-ImageNet   & $\leq 10{,}000$ & $\leq 10{,}000$ & $0.810\ [0.792, 0.828]$ \\
    ResNet-Scratch    & $> 32{,}250$    & $\leq 10{,}000$  & $0.675\ [0.639, 0.715]$ \\
    \bottomrule
    \end{tabular}
\end{table}

\begin{table}[t]
    \centering
    \caption{Minimum longitudinal training set size for reasonable performance (fine-tuned encoder,
             MLP head). Same crossing-point definition and thresholds as
             Table~\ref{tab:min_training_size_frozen}.}
    \label{tab:min_training_size_finetuned}
    \begin{tabular}{lccc}
    \toprule
    \textbf{Model} 
        & \makecell{\textbf{Min }$n$\\(CI $\geq 0.75$)}
        & \makecell{\textbf{Min }$n$\\(CI $\geq 95\%$ peak)}
        & {Peak C-index} [95\% CI] \\
    \midrule
    \multicolumn{4}{l}{\textit{SSL models}} \\
    SimCLR-NAKO        & $\leq 500$    & $\leq 10{,}000$ & $0.933\ [0.932, 0.935]$ \\
    SimCLR-ImageNet-NAKO   & $\leq 500$    & $\leq 1{,}000$ & $0.931\ [0.929, 0.932]$ \\
    MAE-NAKO        &                  $\leq500 $ &                  $\leq 10{,}000 $& $ 0.918 \ [0.917, 0.920]$ \\
    MAE-ImageNet-NAKO     &   $\leq100$ &$\leq 1{,}000$ & $0.928 \ [0.927, 0.929]$ \\

    \midrule
    \multicolumn{4}{l}{\textit{Foundation models}} \\
    DINOv2-LVD         & $\leq 10{,}000$ & $\leq 10{,}000$ & $0.911\ [0.909, 0.913]$ \\
    DINOv2-NAKO        & $\leq 10{,}000$ & $\leq 10{,}000$ & $0.908\ [0.906, 0.911]$ \\
    RETFound            & $\leq 500$      & $\leq 1{,}000$  & $0.930\ [0.929, 0.932]$ \\
    \midrule
    \multicolumn{4}{l}{\textit{Supervised baselines}} \\
    ResNet-ImageNet  & $\leq 500$    & $\leq 10{,}000$ & $0.932\ [0.930, 0.933]$ \\
    ResNet-Scratch    & $\leq 10{,}000$ & $\leq 10{,}000$ & $0.927\ [0.925, 0.928]$ \\
    \bottomrule
    \end{tabular}
\end{table}

\subsection{The pre-training objective matters more than dataset size or domain}
\label{sec:objective}

Having found that pre-training helps most in the frozen, low-data regime, we asked what property of the pre-training determines how well a frozen encoder transfers: the scale of the pre-training corpus, how closely its domain matches the target task, or the self-supervision objective itself. In the decomposition, corpus scale is captured by the model family, since within a given objective the foundation models were pre-trained on far larger corpora than the in-house encoders (for example, RETFound versus MAE-NAKO).

We found that models with the same pre-training objective performed similarly, whether pre-trained on a large or a modest data sets. RETFound and MAE-ImageNet-NAKO reached almost the same peak ($0.870$ and $0.876$; Table~\ref{tab:peak_performance}), even though RETFound was pre-trained on roughly three times as many images with a model more than twice the size, and MAE-NAKO trailed only modestly ($0.838$). The two SimCLR encoders clustered together ($0.920$ and $0.913$), as did the two DINOv2 variants ($0.916$ and $0.912$). Within each objective, the in-house encoders matched or came close to the larger or more heavily pre-trained one, suggesting that corpus size mattered less than the objective.

To test this formally, we decomposed the variance in performance across the factors that distinguish the encoders. We modelled discrimination (C-index) and
calibration (IBS) as smooth functions of training set size with generalised additive models (GAMs), averaging the five training runs within each model--size--head--protocol combination, and compared nested models by AIC to ask how much each factor explained (see Methods). Adding the pre-training objective to a
baseline of size, protocol, and head led to improved model fit, raising the deviance explained from $0.375$ to $0.798$ for C-Index and from $0.640$ to $0.901$ for IBS, and improving AIC by $203$ and $224$ respectively. Adding the model family or the pre-training domain on top of the objective did not improve the deviance explained or the AIC for either metric (Table~\ref{tab:aic_model_comparison}). The objective thus accounted for essentially all of the explainable variation between encoders; family and domain added almost nothing beyond it.

\begin{figure}[t!]
\centering
\includegraphics{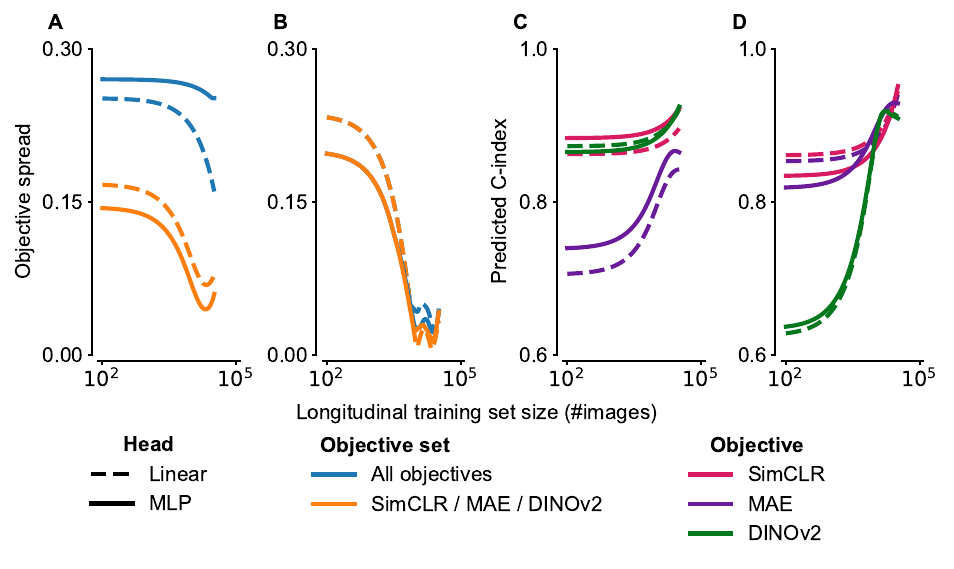}
\caption{The pre-training objective determines performance, specifically in the
frozen and low-data regime (C-index shown; see Appendix~\ref{app:ibs} for
IBS). 
Spread in predicted C-index across objectives (gap between best
and worst performing objectives) in the (\textbf{A}) frozen and (\textbf{B}) fine-tuned encoder protocols as a function of training size and head. The spread is
largest at small training sizes and collapses under fine-tuning as data grows,
showing that the objective matters most when the encoder is frozen, and labels
are few. Predicted discrimination for the three
self-supervision objectives as a function of training set size, under the (\textbf{C}) frozen and (\textbf{D}) fine-tuned protocols and for the linear and MLP survival heads (GAM fitted on seed-averaged data). SimCLR is consistently strong; DINOv2 is strong when frozen
but weakest when fine-tuned at small sample sizes, recovering only with large
training sets; MAE lags when frozen but is stable under fine-tuning.}
\label{fig:objective}
\end{figure}

The objective's influence was most visible in the frozen regime. Stratifying the decomposition by protocol and head, objective explained the bulk of frozen
performance ($\Delta\mathrm{AIC} \approx -112$ for C-Index, raising the proportion of deviance explained from $\sim\!0.167$ to $\sim\!0.897$), but far less under
fine-tuning ($\Delta\mathrm{AIC} \approx -28$ for the linear head and $\approx -3$ for the MLP head), where models of different objectives converge (Supplementary Table~\ref{tab:regime_decomposition}). Its influence also grew as data became scarcer: the spread in predicted C-index across objectives was largest at the smallest training sizes and narrowed as data increased, sharply so under fine-tuning (Figure~\ref{fig:objective}A-B). The objective therefore governs performance specifically in the frozen, low-data
regime -- the same regime in which pre-training itself is most valuable.

Among the three self-supervision objectives, one was stable across regimes while the other two swapped rank between frozen and fine-tuned use (Figure~\ref{fig:objective}C, D). Contrastive (SimCLR) encoders were consistently strong across all training sizes and both protocols. The other two traded places: self-distillation (DINOv2) was strong when frozen but the weakest when fine-tuned at small sample sizes, recovering only with large training sets, whereas masked-autoencoding (MAE) lagged when frozen but was stable and competitive under fine-tuning. The objective differences that persisted under fine-tuning were therefore driven largely by DINOv2's poor data efficiency when adapted, consistent with self-distillation representations being best used off the shelf. Both metrics gave the same picture, with IBS effects smaller in magnitude (Appendix~\ref{app:ibs}).

\begin{table}[t!]
\centering
\caption{Comparison of nested GAMs for predicted C-index and IBS across baseline, objective, family, and in-domain model specifications, reporting degrees of freedom (df), AIC, and proportion of deviance explained.}
\label{tab:aic_model_comparison}
\begin{tabular}{llccc}
\toprule
Metric & Model & df & AIC $\downarrow$ & \makecell{Deviance\\explained $\uparrow$}\\
\midrule
\multirow{4}{*}{C-index}
 &  baseline           & 12.5 & $-515.66$          & 0.375 \\
 & + objective    & 53.0 & $\mathbf{-719.47}$ & $\mathbf{0.798}$ \\
 & + family   & 59.2 & $-700.30$          & 0.792 \\
 & + in-domain & 63.1 & $-696.43$          & 0.795 \\
\midrule
\multirow{4}{*}{IBS}
 & baseline           & 15.7 & $-1694.15$          & 0.640 \\
 & + objective    & 66.7 & $\mathbf{-1918.61}$ & $\mathbf{0.901}$ \\
 & + family   & 75.2 & $-1895.06$          & 0.899 \\
 & + in-domain & 75.2 & $-1888.46$          & 0.896 \\
\bottomrule
\end{tabular}
\end{table}

\section{Discussion}

We compared self-supervised, foundation models, and supervised encoders for retinal disease progression modelling across a wide range of labelled dataset sizes and two evaluation protocols. The benefit of pre-training was concentrated in the frozen, low-data regime that characterises most longitudinal cohorts and
largely disappeared once enough longitudinal data for fine-tuning became available. The self-supervision objective, not the scale of the pre-training corpus or domain match, accounted for essentially all of the explainable difference between encoders, while the model family and the pre-training domain added almost nothing once the objective was known. Interestingly, scale isn't decisive: MAE-ImageNet-NAKO, trained on a modest cross-sectional cohort, matched RETFound's performance despite far fewer images and parameters -- both share a masked-autoencoding objective. Nor is domain adaptation automatically beneficial: LoRA-adapting DINOv2 to fundus images left its performance unchanged regardless of whether the backbone was frozen or fine-tuned. For SimCLR, neither the natural-image initialisation nor the in-domain data was decisive: SimCLR-ImageNet-NAKO, initialised from ImageNet weights before contrastive pre-training on NAKO, performed almost identically to SimCLR-NAKO, trained without ImageNet initialisation.

The value of pre-training depended strongly on how the encoder was used. Frozen, pre-trained representations reached reasonable discrimination from a few hundred labelled samples, while a randomly initialised network did not reach comparable performance even with tens of thousands. Once fine-tuning was feasible, this advantage largely vanished: a standard ImageNet-initialised ResNet became competitive with every foundation model at far lower computational cost. The objectives themselves suited different regimes. Contrastive (SimCLR) encoders were strong throughout, while self-distillation and masked-autoencoding traded places -- DINOv2 was strong frozen but declined when fine-tuned, recovering only with large training sets, whereas MAE-based encoders lagged when frozen but were stable under fine-tuning. DINOv2's poor data efficiency when adapted is consistent with prior evidence that its representations are best used off the shelf~\citep{oquab2024dinov}, making it a strong frozen feature extractor but a poor candidate for end-to-end adaptation. The relative weakness of the MAE-based encoders when frozen may reflect that masked-image reconstruction optimises for pixel-level detail rather than the semantic, disease-relevant structure a frozen shallow head can exploit; fine-tuning, which reshapes the representation, closes much of this gap.

These findings map directly onto the conditions under which progression models are usually built. A clinic or research group assembling a longitudinal cohort typically has a few hundred to a few thousand labelled examples, limited compute, and no domain-specific foundation model of its own. This is precisely the frozen, low-data regime in which pre-training helps most. Our results suggest a simple and inexpensive recipe for this setting: take a frozen encoder pre-trained by self-supervision, whether an off-the-shelf model or one trained on an available cross-sectional cohort, and fit a lightweight survival head on the longitudinal labels. This reaches clinically reasonable discrimination from a few hundred samples without the labelled data, compute, or engineering that full fine-tuning demands, and without a large or domain-matched pre-training corpus. Because abundant cross-sectional imaging exists for many diseases while longitudinal cohorts remain scarce, the same recipe could extend progression modelling to settings where it is currently impractical.

Our study has limitations. We evaluated a single disease (AMD) in a single imaging modality (colour fundus photography), so whether the regime-dependence we observe generalises to other progression tasks or to modalities such as OCT remains open, though the mechanism is not specific to AMD. Also, our absolute performance is not directly comparable to prior AREDS progression models such as those of \citet{yan2020deep} and \citet{peng2020predicting}, which use different inputs, label definitions, and evaluation protocols; our contribution is the comparison across pre-training strategies and data regimes rather than a new state of the art on AREDS.

In summary, we found that the right model choice depends on the available resources. When only frozen feature extraction is practical, which is the realistic setting for small longitudinal cohorts, self-supervised and foundation model encoders reach usable discrimination from the least data, with the pre-training objective as the main determinant of performance. When fine-tuning is feasible, model choice matters far less, except that DINOv2 should not be fine-tuned, and a standard ImageNet-initialised ResNet becomes fully competitive at substantially lower cost. For the data-scarce, compute-constrained conditions typical of longitudinal medical imaging, a self-supervised encoder trained on abundant cross-sectional data is a practical and effective way to build disease progression models.

\clearpage
\newpage

\section*{Acknowledgments}
This project was supported by the Hertie Foundation and by the Deutsche Forschungsgemeinschaft under Germany's Excellence Strategy with the Excellence Cluster 2064 ``Machine Learning — New Perspectives for Science'', project number 390727645. PB is a member of the Else Kröner Medical Scientist Kolleg ``ClinbrAIn: Artificial Intelligence for Clinical Brain Research''. The authors thank the International Max Planck Research School for Intelligent Systems (IMPRS-IS) for supporting IVN and SM. The authors further thank the investigators who contributed to the NAKO ophthalmological dataset \cite{Roa2026.05.04.26352019}: Klaus Berger, Caroline Brandl, Titus Brinker, Anne Elbrecht, Gerd Geerling, Halina Greiser, Carsten Grohmann, Kathrin Günther, Iris Heid, Andre Karch, Thomas Keil, Jessica Krepel, Michael Leitzmann, Claudia Meinke-Franze, Annette Peters, Sabine Schipf, Matthias Schulz, Alexander Schuster, Stefan Willich, Robert Finger, Alexandra Schweig, Martin Leitritz, and Marius Ueffing. This project was conducted with data (Application No. NAKO-590) from the German National Cohort (NAKO) (www.nako.de). The NAKO is funded by the Federal Ministry of Research, Technology and Space (BMFTR) [project funding reference numbers: 01ER1301A/B/C,
01ER1511D, 01ER1801A/B/C/D and 01ER2301A/B/C], federal states of Germany and the Helmholtz Association, the participating universities, and the institutes of the Leibniz Association. We thank all participants who took part in the NAKO study and the staff of this research initiative.

\section*{Author contributions}
\textbf{Conceptualization:} PB, SM, IVN. \textbf{Data curation:} JG, IVN. \textbf{Formal analysis:} PB, JG, IVN. \textbf{Funding acquisition:} PB. \textbf{Investigation:} PB, SM, IVN. \textbf{Methodology:} IVN, JG, SM. \textbf{Software:} JG, IVN. \textbf{Supervision:} PB, SM. \textbf{Visualization:} PB, IVN. \textbf{Writing -- original draft:} IVN, PB. \textbf{Writing -- review \& editing:} SM, JG.

\section*{Data availability}
The AREDS dataset is available from the database of Genotypes and Phenotypes (dbGaP) under accession number phs000001.v3.p1 through the standard dbGaP application process. The NAKO Ophthalmology dataset is available to researchers upon application to the German National Cohort (NAKO) via \url{https://transfer.nako.de}, subject to the cohort's data-use and governance requirements. Neither dataset can be redistributed by the authors. The derived, de-identified model outputs (per-run C-index and IBS values) used for the statistical analyses are provided in the code repository.

\section*{Code availability}
Code for pre-training, survival modelling, and the statistical analyses, together with the derived results used to generate the figures and tables, is available at \url{https://github.com/berenslab/leverage-data}.

\section*{Declaration of Generative AI and AI-assisted technologies in the writing process}

During the preparation of this work the authors used Claude (Anthropic) in order to improve the language, clarity, and structure of the manuscript text as well as Grammarly to improve language. After using this tool, the authors reviewed and edited the content as needed and take full responsibility for the content of the publication.

\clearpage
\newpage
\bibliographystyle{plainnat}
\bibliography{references}

\clearpage
\newpage
\setcounter{section}{0}
\renewcommand{\thesection}{S\arabic{section}}
\renewcommand{\thefigure}{S\arabic{figure}}
\renewcommand{\thetable}{S\arabic{table}}
\renewcommand{\theequation}{S\arabic{equation}}

\section{Methods Extension}
\begin{table}[h!]
    \caption{Summary of Encoders. $^{*}$For DINOv2-NAKO, the trainable-parameter count refers to the LoRA adapters; the underlying ViT-B backbone ($\approx$86M parameters) is frozen during adaptation.}
    \label{tab:models} 
    \centering
    \begin{tabular}{ccccc}
        \toprule
        \# & \makecell{Model\\Name} &\makecell{Model\\Family}&
         {Arch}  & {\makecell{Trainable\\ Params }} \\
\hline
 1 & ResNet-Scratch&Sup.&ResNet18 & $\approx$ 11M \\
&&&\\
2& ResNet-ImageNet & Sup. &ResNet18 & $\approx$ 11M \\
&&&\\
\hline
3&SimCLR NAKO & SSL & ResNet18 &  $\approx$ 11M\\
&&&\\
4 & \makecell{SimCLR-ImageNet\\NAKO} & SSL& ResNet18 &  $\approx$ 11M\\
&&&\\
\hline
5&RETFound&FM&ViT-L & $\approx$ 303M \\
&&&\\
6&MAE-NAKO & SSL&ViT-B & $\approx$ 111M\\
&&&\\
7&MAE-ImageNet-NAKO & SSL&ViT-B & $\approx$ 111M\\
&&&\\
\hline
8&DINOv2 LVD & FM&ViT-B & $\approx$ 86M\\
&&&\\
9&DINOv2 NAKO &FM& ViT-B & $\approx$ 2M$^{*}$\\
\hline
\end{tabular}
\end{table}

\clearpage
\newpage
\section{Additional Figures}
\begin{figure}[h!]
\centering
\includegraphics[scale=1]{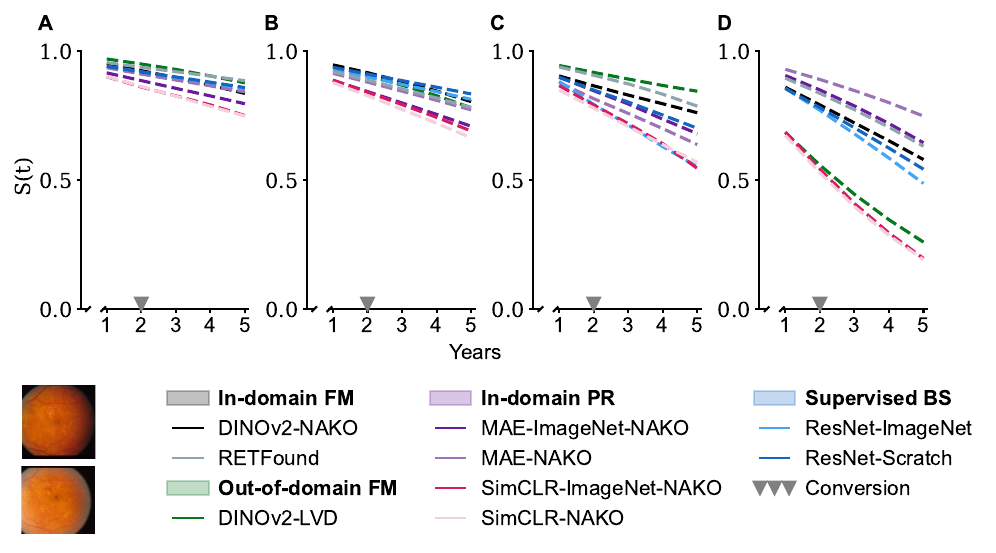}
\caption{Survival curves $S(t)$ reflecting the probability of remaining free from late AMD progression in the next 1--5 years, for one eye, based on the fundus image at year 2 (shown at bottom left), as predicted using different encoder models with a linear output head. The grey triangle marks the time of observed conversion (fundus image shown at bottom right). (\textbf{A, B}) Survival model with frozen encoder, with read-out trained on (\textbf{A}) 1,000 and (\textbf{B}) 32,250  examples. (\textbf{C, D}) Survival model with fine-tuned encoder, trained on 1,000  (\textbf{C}) and 30,000 (\textbf{D}) examples. Colours indicate model family (see legend).}
\label{fig:survival_curves_linear}
\end{figure}

\clearpage
\newpage
\section{Additional Tables}

\begin{table}[h!]
\centering
\caption{Performance at full training size $n = 32{,}250$ and $n = 1{,}000$  (fine-tuned encoder, MLP head). Values are mean $\pm$ SD across five training run repetitions.}
\label{tab:finetune_mlp}
\setlength{\tabcolsep}{6pt}
\begin{tabular}{llccc}
\toprule
\textbf{Model Family} & \textbf{Model Name} &\textbf{ Train Size} &\textbf{ C-index} $\uparrow$& \textbf{IBS }$\downarrow$  \\
\midrule

 \multirow{8}{*}{SSL}& \multirow{2}{*}{SimCLR-ImageNet-NAKO} & {1,000} & 0.893 $\pm$ 0.006 & 0.064 $\pm$ 0.001 \\
 &  & {32,250} & 0.931 $\pm$ 0.002 & 0.051 $\pm$ 0.001 \\
\cmidrule{2-5}

 & \multirow{2}{*}{SimCLR-NAKO} & {1,000} & 0.892 $\pm$ 0.007 & 0.065 $\pm$ 0.002 \\
 &  & {32,250} & 0.932 $\pm$ 0.001 & 0.051 $\pm$ 0.001 \\
 \cmidrule{2-5}
 
 &\multirow{2}{*}{MAE-ImageNet-NAKO} & {1,000} & 0.888 $\pm$ 0.005 & 0.063 $\pm$ 0.000 \\
 &  & {32,250} & 0.929 $\pm$ 0.001 & 0.051 $\pm$ 0.000 \\
 \cmidrule{2-5}
 
 &\multirow{2}{*}{MAE-NAKO} & {1,000} & 0.865 $\pm$ 0.004 & 0.065 $\pm$ 0.000 \\
 &  & {32,250} & 0.917 $\pm$ 0.002 & 0.055 $\pm$ 0.001 \\
 
\midrule
\multirow{6}{*}{FM} & \multirow{2}{*}{DINOv2-LVD} & {1,000} & 0.683 $\pm$ 0.057 & 0.085 $\pm$ 0.004 \\
 &  & {32,250} & 0.911 $\pm$ 0.003 & 0.057 $\pm$ 0.001 \\
\cmidrule{2-5}

 & \multirow{2}{*}{DINOv2-NAKO} & {1,000} & 0.765 $\pm$ 0.057 & 0.077 $\pm$ 0.005 \\
 &  & {32,250} & 0.908 $\pm$ 0.003 & 0.057 $\pm$ 0.001 \\
\cmidrule{2-5}

 & \multirow{2}{*}{RETFound} & {1,000} & 0.898 $\pm$ 0.014 & 0.063 $\pm$ 0.002 \\
 &  & {32,250} & 0.929 $\pm$ 0.001 & 0.052 $\pm$ 0.001 \\
\midrule
\multirow{4}{*}{Sup.} & \multirow{2}{*}{ResNet-ImageNet} & {1,000} & 0.869 $\pm$ 0.014 & 0.068 $\pm$ 0.003 \\
 &  & {32,250} & 0.932 $\pm$ 0.002 & 0.051 $\pm$ 0.001 \\
\cmidrule{2-5}

 & \multirow{2}{*}{ResNet-Scratch} & {1,000} & 0.788 $\pm$ 0.045 & 0.077 $\pm$ 0.005 \\
 &  & {32,250} & 0.927 $\pm$ 0.002 & 0.052 $\pm$ 0.001 \\
\bottomrule
\end{tabular}
\end{table}

\begin{table}[h!]
\centering
\caption{Performance at full training size $n = 32{,}250$ and $n = 1{,}000$  (frozen encoder, linear head). Values are mean $\pm$ SD across five training run repetitions.}
\label{tab:frozen_linear}
\setlength{\tabcolsep}{6pt}
\begin{tabular}{llccc}
\toprule
\textbf{Model Family} & \textbf{Model Name} &\textbf{ Train Size} &\textbf{ C-index} $\uparrow$& \textbf{IBS }$\downarrow$  \\

\midrule
 \multirow{8}{*}{SSL} & \multirow{2}{*}{SimCLR-ImageNet-NAKO} & {1,000} & 0.877 $\pm$ 0.001 & 0.070 $\pm$ 0.000 \\
 &  & {32,250} & 0.893 $\pm$ 0.000 & 0.062 $\pm$ 0.000 \\
\cmidrule{2-5}

 & \multirow{2}{*}{SimCLR-NAKO} & {1,000} & 0.856 $\pm$ 0.012 & 0.072 $\pm$ 0.001 \\
 &  & {32,250} & 0.888 $\pm$ 0.006 & 0.064 $\pm$ 0.002 \\
 \cmidrule{2-5}

& \multirow{2}{*}{MAE-ImageNet-NAKO} & {1,000} & 0.780 $\pm$ 0.001 & 0.081 $\pm$ 0.000 \\
 &  & {32,250} & 0.830 $\pm$ 0.000 & 0.071 $\pm$ 0.000 \\
\cmidrule{2-5} 

& \multirow{2}{*}{MAE-NAKO} & {1,000} & 0.723 $\pm$ 0.008 & 0.082 $\pm$ 0.000 \\
 &  & {32,250} & 0.824 $\pm$ 0.006 & 0.071 $\pm$ 0.000 \\
\midrule

\multirow{6}{*}{FM} & \multirow{2}{*}{DINOv2-LVD} & {1,000} & 0.893 $\pm$ 0.004 & 0.066 $\pm$ 0.001 \\
 &  & {32,250} & 0.912 $\pm$ 0.002 & 0.060 $\pm$ 0.002 \\
\cmidrule{2-5}

 & \multirow{2}{*}{DINOv2-NAKO} & {1,000} & 0.894 $\pm$ 0.003 & 0.066 $\pm$ 0.002 \\
 &  & {32,250} & 0.913 $\pm$ 0.001 & 0.060 $\pm$ 0.001 \\
\cmidrule{2-5}
 & \multirow{2}{*}{RETFound} & {1,000} & 0.775 $\pm$ 0.007 & 0.081 $\pm$ 0.001 \\
 &  & {32,250} & 0.860 $\pm$ 0.002 & 0.069 $\pm$ 0.001 \\
\midrule
\multirow{4}{*}{Sup.} & \multirow{2}{*}{ResNet-ImageNet} & {1,000} & 0.722 $\pm$ 0.014 & 0.085 $\pm$ 0.002 \\
 &  & {32,250} & 0.770 $\pm$ 0.015 & 0.080 $\pm$ 0.002 \\
\cmidrule{2-5}
 & \multirow{2}{*}{ResNet-Scratch} & {1,000} & 0.642 $\pm$ 0.011 & 0.088 $\pm$ 0.001 \\
 &  & {32,250} & 0.754 $\pm$ 0.022 & 0.079 $\pm$ 0.002 \\
\bottomrule
\end{tabular}
\end{table}
\clearpage
\newpage
\begin{table}[t!]
\centering
\caption{Performance at full training size $n = 32{,}250$ and $n = 1{,}000$  (fine-tuned encoder, linear head). Values are mean $\pm$ SD across five training run repetitions.}
\label{tab:finetune_linear}
\setlength{\tabcolsep}{6pt}
\begin{tabular}{llccc}
\toprule
\textbf{Model Family} & \textbf{Model Name} &\textbf{ Train Size} &\textbf{ C-index} $\uparrow$& \textbf{IBS }$\downarrow$  \\
\midrule

 \multirow{8}{*}{SSL} & \multirow{2}{*}{SimCLR-ImageNet-NAKO}& {1,000} & 0.898 $\pm$ 0.005 & 0.064 $\pm$ 0.001 \\
 &  & {32,250} & 0.930 $\pm$ 0.002 & 0.051 $\pm$ 0.001 \\
\cmidrule{2-5}
 & \multirow{2}{*}{SimCLR-NAKO} & {1,000} & 0.895 $\pm$ 0.005 & 0.064 $\pm$ 0.001 \\
 &  & {32,250} & 0.932 $\pm$ 0.002 & 0.051 $\pm$ 0.001 \\
 
\cmidrule{2-5}
 & \multirow{2}{*} {MAE-ImageNet-NAKO} & {1,000} & 0.894 $\pm$ 0.002 & 0.062 $\pm$ 0.000 \\
 &  & {32,250} & 0.928 $\pm$ 0.002 & 0.050 $\pm$ 0.000 \\
 
 \cmidrule{2-5}
 & \multirow{2}{*} {MAE-NAKO} & {1,000} & 0.869 $\pm$ 0.009 & 0.063 $\pm$ 0.002 \\
 &  & {32,250} & 0.918 $\pm$ 0.002 & 0.054 $\pm$ 0.001 \\
\midrule

\multirow{6}{*}{FM} & \multirow{2}{*}{DINOv2-LVD} & {1,000} & 0.703 $\pm$ 0.044 & 0.085 $\pm$ 0.002 \\
 &  & {32,250} & 0.912 $\pm$ 0.002 & 0.056 $\pm$ 0.000 \\
\cmidrule{2-5}
 & \multirow{2}{*}{DINOv2-NAKO} & {1,000} & 0.703 $\pm$ 0.052 & 0.083 $\pm$ 0.004 \\
 &  & {32,250} & 0.912 $\pm$ 0.002 & 0.056 $\pm$ 0.000 \\
\cmidrule{2-5}
 & \multirow{2}{*}{RETFound} & {1,000} & 0.901 $\pm$ 0.006 & 0.061 $\pm$ 0.002 \\
 &  & {32,250} & 0.932 $\pm$ 0.004 & 0.051 $\pm$ 0.002 \\
\midrule
\multirow{4}{*}{Sup.} & \multirow{2}{*}{ResNet-ImageNet} & {1,000} & 0.868 $\pm$ 0.007 & 0.071 $\pm$ 0.001 \\
 &  & {32,250} & 0.929 $\pm$ 0.002 & 0.051 $\pm$ 0.001 \\
\cmidrule{2-5}
 & \multirow{2}{*}{ResNet-Scratch} & {1,000} & 0.840 $\pm$ 0.005 & 0.073 $\pm$ 0.002 \\
 &  & {32,250} & 0.928 $\pm$ 0.003 & 0.051 $\pm$ 0.001 \\
\bottomrule
\end{tabular}
\end{table}

\clearpage
\newpage

\section{Further Analysis}

\begin{figure}[h!]
\centering
\includegraphics{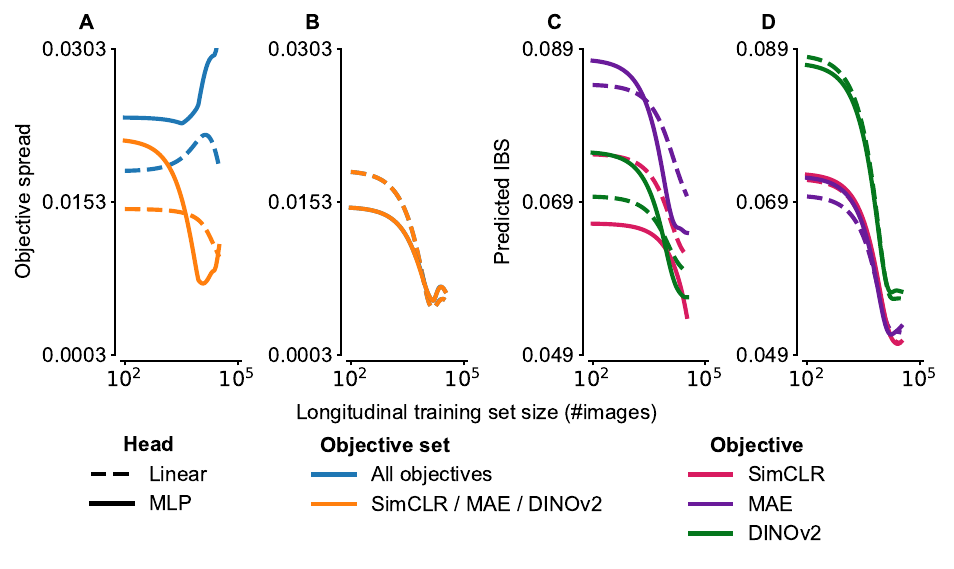}
\caption{
Spread in predicted IBS across objectives in the (\textbf{A}) frozen and (\textbf{B}) fine-tuned encoder protocols as a function of training size and head. Predicted discrimination for the three
self-supervision objectives as a function of training set size, under the (\textbf{C}) frozen and (\textbf{D}) fine-tuned protocols and for the linear and MLP survival heads.}
\label{fig:objective_ibs}
\end{figure}

\label{app:ibs}
\begin{table}[h!]
\centering
\caption{Regime-specific deviance decomposition for the full 5-objective dataset, showing  the proportion of deviance explained in the baseline (Dev-Base), after adding the objective (Dev-Obj), and the corresponding AIC improvement ($\Delta$AIC).}
\label{tab:regime_decomposition}
\begin{tabular}{lrrrrr}
\toprule
Metric & Protocol & Head & Dev-Base  $\uparrow$ & Dev-Obj $\uparrow$& \makecell{$\Delta$AIC $\downarrow$\\(+objective)} \\
\midrule
\multirow{4}{*}{C-index}
 & Fine-tuned & Linear & 0.492 & 0.761 & $-28.27$ \\
 & Fine-tuned & MLP    & 0.516 & 0.651 & $-3.01$ \\
 & Frozen     & Linear & 0.167 & 0.897 & $\mathbf{-111.59}$ \\
 & Frozen     & MLP    & 0.123 & 0.877 & $\mathbf{-104.25}$ \\
\midrule
\multirow{4}{*}{IBS}
 & Fine-tuned & Linear & 0.747 & 0.886 & $-21.77$ \\
 & Fine-tuned & MLP    & 0.755 & 0.857 & $-6.60$ \\
 & Frozen     & Linear & 0.324 & 0.945 & $\mathbf{-130.73}$ \\
 & Frozen     & MLP    & 0.356 & 0.896 & $\mathbf{-91.98}$ \\
\bottomrule
\end{tabular}
\end{table}

\begin{table}[h!]
\centering
\caption{Regime-specific deviance decomposition for the 3-objective (SimCLR/MAE/DINOv2) subset, showing baseline proportion of deviance explained (Dev-Base), deviance explained after adding objective (Dev-Obj), and the corresponding AIC improvement ($\Delta$AIC).}
\label{tab:regime_decomposition_ssl_subset}
\begin{tabular}{lrrrrr}
\toprule
Metric & Protocol & Head & Dev-Base $\uparrow$ & Dev-Obj $\uparrow$ & \makecell{$\Delta$AIC $\downarrow$\\(+objective)} \\
\midrule
\multirow{4}{*}{C-index}
 & Fine-tuned & Linear & 0.445 & 0.830 & $-47.01$ \\
 & Fine-tuned & MLP    & 0.471 & 0.645 & $-11.25$ \\
 & Frozen     & Linear & 0.207 & 0.856 &$\mathbf{-71.69}$\\
 & Frozen     & MLP    & 0.273 & 0.764 & $\mathbf{-44.75}$ \\
\midrule
\multirow{4}{*}{IBS}
 & Fine-tuned & Linear & 0.707 & 0.882 & $-29.25$ \\
 & Fine-tuned & MLP    & 0.715 & 0.837 & $-13.38$ \\
 & Frozen     & Linear & 0.478 & 0.924 & $\mathbf{-76.75}$ \\
 & Frozen     & MLP    & 0.549 & 0.850 & $\mathbf{-42.02}$ \\
\bottomrule
\end{tabular}
\end{table}

\end{document}